\documentclass[twoside,11pt]{article}

\usepackage{jmlr2e}
\usepackage{amsmath,amssymb}
\usepackage{qtree}
\usepackage{subfigure}
\usepackage{algorithm}
\usepackage{algpseudocode}

\ShortHeadings{Reinforced Decision Trees}{}
\firstpageno{1}

\begin{document}

\title{Reinforced Decision Trees}

\author{\name Aur\'elia L\'eon \email aurelia.leon@lip6.fr \\
\addr Sorbonne Universit\'es, UPMC Univ Paris 06, UMR 7606, LIP6, F-75005, Paris, France\\
        \name Ludovic Denoyer \email ludovic.denoyer@lip6.fr\\
        \addr Sorbonne Universit\'es, UPMC Univ Paris 06, UMR 7606, LIP6, F-75005, Paris, France}


\maketitle

\begin{abstract}
In order to speed-up classification models when facing a large number of categories, one usual approach consists in organizing the categories in a particular structure, this structure being then used as a way to speed-up the prediction computation. This is for example the case when using error-correcting codes or even hierarchies of categories. But in the majority of approaches, this structure is chosen \textit{by hand}, or during a preliminary step, and not integrated in the learning process. We propose a new model called Reinforced Decision Tree which simultaneously learns how to organize categories in a tree structure and how to classify any input based on this structure. This approach keeps the advantages of existing techniques (low inference complexity) but allows one to build efficient classifiers in one learning step. The learning algorithm is inspired by reinforcement learning and policy-gradient techniques which allows us to integrate the two steps (building the tree, and learning the classifier) in one single algorithm.
\end{abstract}


\section{Introduction}

The complexity of classification models is usually highly related, and typically linear w.r.t the number of possible categories denoted $C$. When facing problems with a very large number of classes, like text classification in large ontologies, object recognition or word prediction in deep learning language models, this becomes a critical point making classification methods inefficient in term of inference complexity. There is thus a need to develop new methods able to predict in large output spaces at a low cost.

Several methods have been recently developed for reducing the classification speed. They are based on the idea of using a \textit{structure} that organizes the possible outputs, and allows one to reduce the inference complexity. Two main families of approaches have been proposed: (i) error-correcting codes approaches \citep{Dietterich1995a,Schapire1997,Cisse2012} that associate a short code to each category; the classification becomes a code prediction problem which can be achieved faster by predicting each element of the code. (ii)  The second family is hierarchical methods \citep{Bengio2010, Liu2013, Weston2013} where the possible categories are leaves of a tree. In that case, an output is predicted by choosing a path in the tree like in decision trees. These two methods typically involve a prediction complexity of $\mathcal{O}(\log C)$ allowing a great speed-up.

But these approaches suffer from one major drawback: the structure used for prediction (error correcting codes, or tree) is usually built during a preliminary step - before learning the classifier - following hand-made heuristics, typically by using clustering algorithms or Huffman codes \citep{mikolov}. The problem is that the quality of the obtained structure greatly influences the quality of the final classifier, and this step is thus a critical step.

In this paper, we propose the \textit{Reinforced Decision Tree (RDT)} model which is able to simultaneously learn how to organize categories in a hierarchy and learn the corresponding classifier in a single step. The obtained system is a high speed and efficient predictive model where the category structure has been fitted for the particular task to solve, while it is built \textit{by hand} in existing approaches. The main idea of RDT is to consider the classification problem as a \textbf{sequential decision process} where a learned policy guides any input $x$ in a tree structure from the root to one leaf. The learning algorithm is inspired from policy-gradient methods \citep{reinforce} which allows us to act on both the way an input $x$ falls into the tree, but also on the categories associated with the leaves of this tree. The difference with the classical Reinforcement Learning context is that the feedback provided to the system is a derivable loss function as proposed in \citep{dsnn} which gives more information than a reward signal, and allows fast learning of the parameters.

The contributions are:
\begin{itemize}
\item A novel model able to simultaneously discover a relevant hierarchy of categories and the associated classification model in one step. 
\item A gradient-based learning method based on injecting a derivable loss function in policy gradient algorithms.
\item A first set of preliminary experiments over toy datasets allowing to better understand the properties of such an approach.
\end{itemize}

\section{Notations and Model}

We consider the multi-class classification problem where each input $x \in \mathbb{R}^n$ has to be associated with one of the $C$ possible categories\footnote{The model naturally handles multi-label classification problems which will not be detailed here for sake of simplicity}. Let us denote $y$ the label of $x$, $y \in \mathbb{R}^C$ such that $y_i=1$ if $x$ belongs to class $i$ and $y_i=-1$ elsewhere. We will denote $\{(x^1,y^1), ...,(x^N,y^N)\}$ the set of $N$ training examples.

\subsection{Reinforced Decision Trees Architecture}

\begin{figure}[ht]
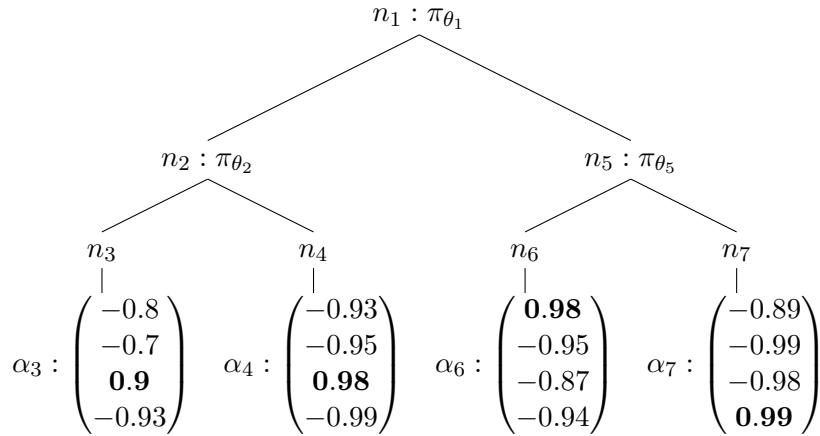

\begin{center}
\Tree[.$n_1:\pi_{\theta_1}$			
            	[.$n_2:\pi_{\theta_2}$
                	[.$n_3$ $\alpha_3:\begin{pmatrix}-0.8\cr -0.7\cr  \mathbf{0.9}\cr  -0.93\end{pmatrix}$
                    ]
                    [.$n_4$ $\alpha_4:\begin{pmatrix}-0.93\cr -0.95\cr \mathbf{0.98}\cr -0.99\end{pmatrix}$
                    ]
                ]            
                [.$n_5:\pi_{\theta_5}$                    
                        [.$n_6$ $\alpha_6:\begin{pmatrix}\mathbf{0.98}\cr -0.95\cr -0.87\cr -0.94\end{pmatrix}$
                        ] 
                        [.$n_7$ $\alpha_7:\begin{pmatrix}-0.89\cr -0.99\cr -0.98\cr \mathbf{0.99}\end{pmatrix}$
                        ]
                ]
       ]
\end{center}
\caption{An example of RDT. The $\pi_\theta$ functions and the $\alpha$ values have been learned in one integrated step following a policy-gradient based method. Bold values correspond to predicted categories at the leaves level. Note that different leaves can predict the same category. The inference speedup comes from the fact that, in this case, the score of the 4 categories is computed by taking at most 2 decisions - for example $\pi_{\theta_1}, \pi_{\theta_2}$, and then returning $\alpha_3$ as a prediction. $\theta$ and $\alpha$ are learned.}
\label{fig:rdt}
\end{figure}

The \textit{Reinforced Decision Tree (RDT)} architecture shares common points with decision trees. Let us denote $\mathcal{T}_{\theta,\alpha}$ such a tree with parameters $\theta$ and $\alpha$ that will be defined in the later:  
\begin{itemize}
\item The tree is composed of a set of nodes denoted $nodes(\mathcal{T}_{\theta,\alpha}) = \{n_1,...,n_T\}$ where $T$ is the number of nodes of the tree
\item $n_1$ is the root of the tree.
\item $parent(n_i)=n_j$ means that node $n_j$ is the parent of $n_i$. Each node but $n_1$ has only one parent, $n_1$ has no parent since it is the root of the tree.
\item $leaf(n_i)=true$ if and only if $n_i$ is a leaf of the tree. 
\end{itemize}
Note that we do not have any constraint concerning the number of leaves or the topology of the tree. Each node of the tree is associated with its own set of parameters:
\begin{itemize}
\item A node $n_i$ is associated with a set of parameters denoted $\theta_i$ if $n_i$ is an internal node i.e $leaf(n_i)=false$. 
\item A node $n_i$ is associated with a set of parameters denoted $\alpha_i \in \mathbb{R}^C$ if $n_i$ is a leaf of the tree.
\end{itemize}
Parameters $\theta=\{\theta_i\}$ are the parameters of the policy that will guide an input $x$ from the root node to one of the leaf of the tree. Parameters $\alpha_i$ correspond to the prediction that will be produced when an input reached the leaf $n_i$. Our architecture is thus very close to classical decision trees, the \textbf{major difference} being that the prediction associated with each leaf is a set of parameters that will be learned during the training process, allowing the model to choose how to match the leaves of the tree with the categories. The model will thus be able to simultaneously learn which path an input has to follow based on the $\theta_i$ parameters, but also how to organize the categories in the tree based on the $\alpha_i$ parameters. An example of RDT is given in Figure \ref{fig:rdt}.

\subsection{Inference Process}
Let us denote $H$ a trajectory, $H=(n_{(1)},...,n_{(t)})$ where $n_{(i)} \in nodes(\mathcal{T}_{\theta,\alpha})$ and $(i)$ is the index of the i-th node of the trajectory. $H$ is thus a sequence of nodes where $n_{(1)}=n_1$, $\forall i>1, n_{(i-1)}=parent(n_{(i)})$ and $leaf(n_{(t)})=true$.

Each internal node $n_i$ is associated with a function (or policy) $\pi_{\theta_i}$ which role is to compute the probability that a given input $x$ will fall in one of the children of $n_i$. $\pi_{\theta_i}$ is defined as:
\begin{equation}
\begin{aligned}
\pi_{\theta_i}&: \mathbb{R}^n \times nodes(\mathcal{T_{\theta,\alpha}}) \rightarrow \mathbb{R}  \\
\pi_{\theta_i}&(x,n) = P(n|n_i,x)
\end{aligned}
\end{equation}
with $P(n|n_i,x)=0$ if $n \notin children(n_i)$. In other words, $\pi_{\theta_i}(x,n)$ is the probability that $x$ moves from $n_i$ to $n$. Note that the different between RDT and DT is that the decision taken at each node is stochastic and not deterministic which will allow us to use gradient-based learning algorithms. The probability of a trajectory $H=(n_{(1)},...,n_{(t)})$ given an input $x$ can be written as:
\begin{equation}
P(H|x) = \prod\limits_{i=1}^{t-1} \pi_{\theta_{(i)}}(x,n_{(i+1)})
\end{equation}

Once a trajectory has been sampled, the prediction produced by the model depends on the leaf $n_{(t)}$ reached by $x$. The model directly outputs $\alpha_{(t)}$ as a prediction, $\alpha_{(t)}$ being a vector in $\mathbb{R}^C$ (one score per class) as explained before. Note that the model produces one score for each possible category, but the inference complexity of this step is $\mathcal{O}(1)$ since it just corresponds to returning the value $\alpha_{(t)}$.

The complete inference process is described in Algorithm \ref{algo}:
\begin{enumerate}
\item Sample a trajectory $H=(n_{(1)},...,n_{(t)})$ given $x$, by sequentially using the policies $\pi_{\theta_{(1)}},...,\pi_{\theta_{(t-1)}}$
\item Returns the predicted output $\alpha_{(t)}$
\end{enumerate}

\subsection{Learning Process}

\begin{algorithm}[t]
\caption{Stochastic Gradient RDT Learning Procedure}\label{alg_inference}
\small{
\begin{algorithmic}[1]
\Procedure{Learning}{$(x^1,y^1),...,(x^N,y^N)$}\Comment{the training set}
\State $\epsilon$ is the learning rate
\State $\alpha \approx random$
\State $\theta \approx random$
\Repeat
\State $i \approx uniform(1,N)$
\State Sample $H=(n_{(1)},...,n_{(t)})$ using the $\pi_\theta(x^i)$ functions
\State $\alpha_{(t)} \gets \alpha_{(t)} - \epsilon \nabla_{\alpha} \Delta(\alpha_{(t)},y^i)$
\For{$k \in [1..t-1]$}
\State $\theta_{(k)} \gets \theta_{(k)}  - \epsilon \left( \nabla_{\theta} \log \pi_{\theta_{(k)}}(x^i) \right) \Delta(\alpha_{(t)},y^i)$
\EndFor
\Until{Convergence}
\State \textbf{return} $\mathcal{T}_{\theta,\alpha}$
\EndProcedure
\end{algorithmic}
}
\label{algo}
\end{algorithm}

The goal of the learning procedure is to simultaneously learn both the \textit{policy functions} $\pi_{\theta_i}$ and the \textit{output parameters} $\alpha_i$ in order to minimize a given learning loss denoted $\Delta$ which corresponds here to a classification loss (e.g square loss or hinge loss)\footnote{Note that our approach is not restricted to classification and can be used for regression for example, see Section \ref{exp}.}

Our learning algorithm is based on an extension of \textbf{policy gradient techniques} inspired from the Reinforcement Learning literature  and similar to \cite{dsnn}. More precisely, our learning method is close to the methods proposed in \cite{reinforce} with the difference that, instead of considering a reward signal which is usual in reinforcement learning, we consider a loss function $\Delta$. This function computes the quality of the system, providing a richer feedback information than simple rewards since it can be derivated, and thus gives the direction in which parameters have to be updated.\hfill~\linebreak

 The performance of our system is denoted $J(\theta,\alpha)$:
\begin{equation}
J(\theta, \alpha) = E_{P_\theta(x,H,y)}[\Delta(F_\alpha(x,H),y)]
\end{equation}
where $F_\alpha(x,H)$ is the prediction made following trajectory $H$ - i.e the sequence of nodes chosen by the $\pi$-functions. The optimization of $J$ can be made by gradient-descent techniques and we need to compute the gradient of $J$: 
\begin{equation}
\begin{aligned}
\nabla_{\theta,\alpha} J(\theta,\alpha) &= \int \nabla_{\theta,\alpha} \left( P_\theta(H|x) \Delta(F_\alpha(x,H),y) \right) P(x,y) dH dx dy \\
\end{aligned}
\end{equation}

This gradient can be simplified such that:
\begin{equation}
\begin{aligned}
\nabla_{\theta,\alpha} J(\theta,\alpha)& = \int \nabla_{\theta,\alpha} \left( P_\theta(H|x) \right) \Delta(F_\alpha(x,H),y) P(x,y) dH dx dy \\
&+ \int  P_\theta(H|x)  \nabla_{\theta,\gamma} \Delta(F_\alpha(x,H),y) P(x,y) dH dx dy\\
&=  \int P_\theta(H|x) \nabla_{\theta,\alpha} \left( log P_\theta(H|x) \right) \Delta(F_\alpha(x,H),y) P(x,y) dH dx dy \\
&+ \int  P_\theta(H|x)  \nabla_{\theta,\alpha} \Delta(F_\alpha(x,H),y) P(x,y) dH dx dy\\
\end{aligned}
\label{eq:eq}
\end{equation}

Using the Monte Carlo approximation of this expectation by taking $M$ trail histories over the $N$ training examples, and given that $\Delta(F_\alpha(x^i,H),y) = \Delta(\alpha_{(t)},y)$, we obtain:
\begin{equation}
\nabla_{\theta,\alpha} J(\theta,\alpha) = \frac{1}{N}\frac{1}{M}\sum\limits_{i=1}^{N} \sum\limits_{k=1}^{M}   \left[ \sum\limits_{j=1}^{t-1} \left( \nabla_{\theta,\alpha} \left( \log \pi_{\theta_{(j)}}(x^i) \right) \Delta(\alpha_{(t)},y) \right) +  \nabla_{\theta,\alpha} \Delta(\alpha_{(t)},y) \right]
\end{equation}

Intuitively, the gradient is composed of two terms:
\begin{itemize}
\item The first term aims at penalizing trajectories with high loss - and thus encouraging to find trajectories with low loss.  The first term only acts on the $\theta$ parameters, modifying the probabilities computed at the internal nodes levels.
\item The second term is the gradient computed over the final loss and concerns the $\alpha$ values corresponding to the leaf node where the input $x$ arrives at the end of the process. It thus changes the $\alpha$ used for prediction in order to capture the category of $x$. This gradient term is responsible about how to allocate the categories in the leaves of the tree.
\end{itemize}

The learning algorithm in its stochastic gradient variant is described in Algorithm \ref{algo}.

\subsection{Discussion}

\paragraph{Inference Complexity: }

The complexity of the inference process is linear with the depth of the tree. Typically, in a multi-class classification problem, the depth of the tree will be proportional to $\log C$, resulting in a very high speed inference process similar to the one obtained for example using Hierarchical SoftMax modules. 

\paragraph{Learning Complexity: }

The policy gradient algorithm developed in this paper is an iterative gradient-based method. Each learning iteration complexity is $\mathcal{O}(N \log C)$ but the number of needed iterations is not known. Moreover, the optimization problem is clearly not a convex problem, and the system can be stuck in a local minimum. As explained in Section \ref{exp}, a way to avoid problematic local minimum is to choose a number of leaves which is higher than the number of categories, giving more freedom degrees to the model.


\paragraph{Using RDT for complex problems}

Different functions topologies can be used for $\pi$. In the following we have used simple linear functions, but more sophisticated ones can be tested like neural networks. Moreover, in our model, there is no constraints upon the $\alpha$ parameters, nor about the loss function $\Delta$ which only has to be derivable. It thus means that our model can also be used for other tasks like multivariate regression, or even for producing continuous outputs at a low price. In that case, RDTs act as discretization processes where the objective of the task and the discretization are made simultaneously.

\section{Preliminary Experiments}
\label{exp}

In this Section, we provide a set of experiments that have been made on toy-datasets to better understand the ability of RDT to perform good predictions and to discover a relevant hierarchy\footnote{Experiments on real-world datasets are currently made}. Our model has been compared to the same model but where the categories associated with the leaves (the $\alpha_i$ values) have been chosen randomly, each leaf being associated to one possible category - i.e each vector $\alpha_i$ is full of $-1$ with one $1$-value at a random position - this model is called \textit{Random Tree}. In that case, the $\alpha$-parameters are not updated during the gradient descent. Hyper-parameters (learning rate and number of iterations) have been tuned by cross-validation, and the results have been averaged on five different runs.  We also made comparisons with a linear SVM (one versus all) and decision trees\footnote{using the implementation of sklearn}. These preliminary experiments aim at showing the ability of our integrated approach to learn how to associate categories to leaves of the tree. 

\begin{figure}
        \begin{minipage}{.50\linewidth}
\small{
                \begin{tabular}{|c|c|c||c|p{1.8cm}|} \hline
  \multicolumn{3}{|c|}{}& \multicolumn{2}{|c|}{Accuracy$\pm$Variance}  \\
  \hline
  $W$ & $D$ & $L$ & RDT  & Random Trees \\
  2& 3 & 8 & 0.46 $\pm$ 0.01 & 0.18 $\pm$ 0.01 \\
  2&4&16 & 0.70 $\pm$ 0.06 & 0.25 $\pm$ 0.02 \\
  2&5&32& 0.83 $\pm$ 0.04 & 0.26 $\pm$ 0.06 \\
  3&2&9& 0.51 $\pm$ 0.01 & 0.25 $\pm$ 0.01 \\
  3&3&27& 0.75 $\pm$ 0.04 & 0.24 $\pm$ 0.02 \\
\hline \hline \hline
\multicolumn{4}{|c|}{Acc. of linear SVM : }& 0.50$\pm$ 0.01 \\ \hline
\multicolumn{4}{|c|}{DT (depth=5) : }& 0.46$\pm$ 0.04 \\ \hline
\multicolumn{4}{|c|}{DT (depth=10) : }& 0.80$\pm$ 0.01 \\ \hline
\multicolumn{4}{|c|}{DT (depth=50) : }& 0.79$\pm$ 0.01 \\ \hline
\end{tabular}
}
        \end{minipage} 
\begin{minipage}{0.50\linewidth}            

\includegraphics[width=1.0\linewidth]{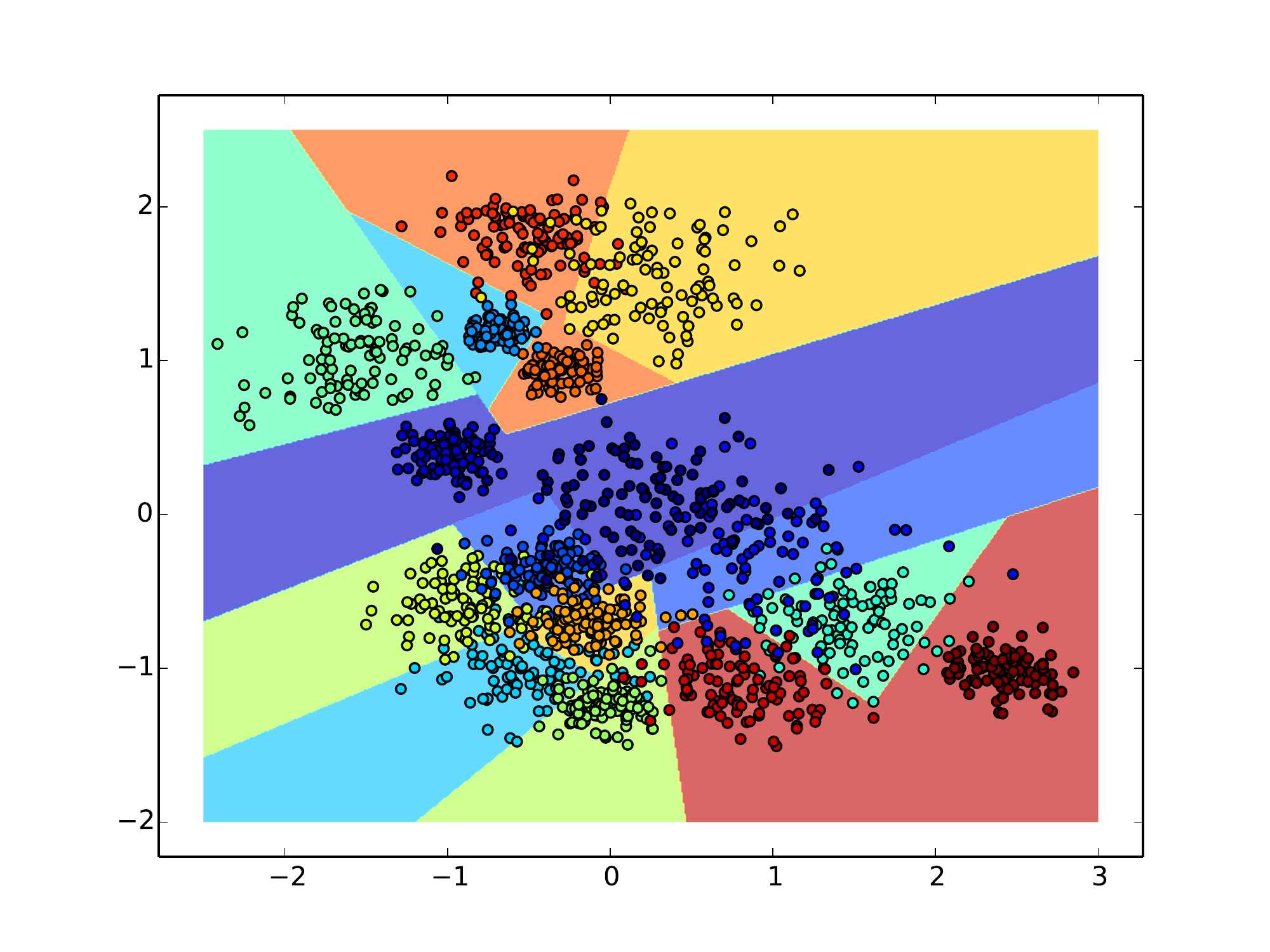}
        \end{minipage}
       
        \caption{Performance (left) on 16 categories and (right) corresponding decision frontiers of RDT: $W$ is the width of the tree (i.e the number of children per node), $D$ is the depth and $L$ is the resulting number of leaves.}
        \label{res2}
\end{figure}

\begin{figure}
        \begin{minipage}{.45\linewidth}
\small{
                \begin{tabular}{|c|c|c||c|p{1.8cm}|} \hline
  \multicolumn{3}{|c|}{}& \multicolumn{2}{|c|}{Accuracy$\pm$Variance}  \\
  \hline
  $W$ & $D$ & $L$ & RDT  & Random Trees \\
  2&5&32& 0.71 $\pm$ 0.04 & 0.10 $\pm$ 0.02 \\
  2&6&64 & 0.84 $\pm$ 0.01 & 0.11 $\pm$ 0.03 \\
  3&3&27& 0.58 $\pm$ 0.04 & 0.12 $\pm$ 0.04 \\
  3&4&81& 0.79 $\pm$ 0.04 & 0.14 $\pm$ 0.06 \\
\hline \hline \hline
\multicolumn{4}{|c|}{Acc. of linear SVM : }& 0.54$\pm$ 0.01 \\ \hline
\multicolumn{4}{|c|}{DT (depth=5) : }& 0.77$\pm$ 0.03 \\ \hline
\multicolumn{4}{|c|}{DT (depth=10) : }& 0.88$\pm$ 0.02 \\ \hline
\multicolumn{4}{|c|}{DT (depth=50) : }& 0.86$\pm$ 0.01 \\ \hline
\end{tabular}
}        
        \end{minipage} 
\begin{minipage}{0.55\linewidth}            
\includegraphics[width=1.0\linewidth]{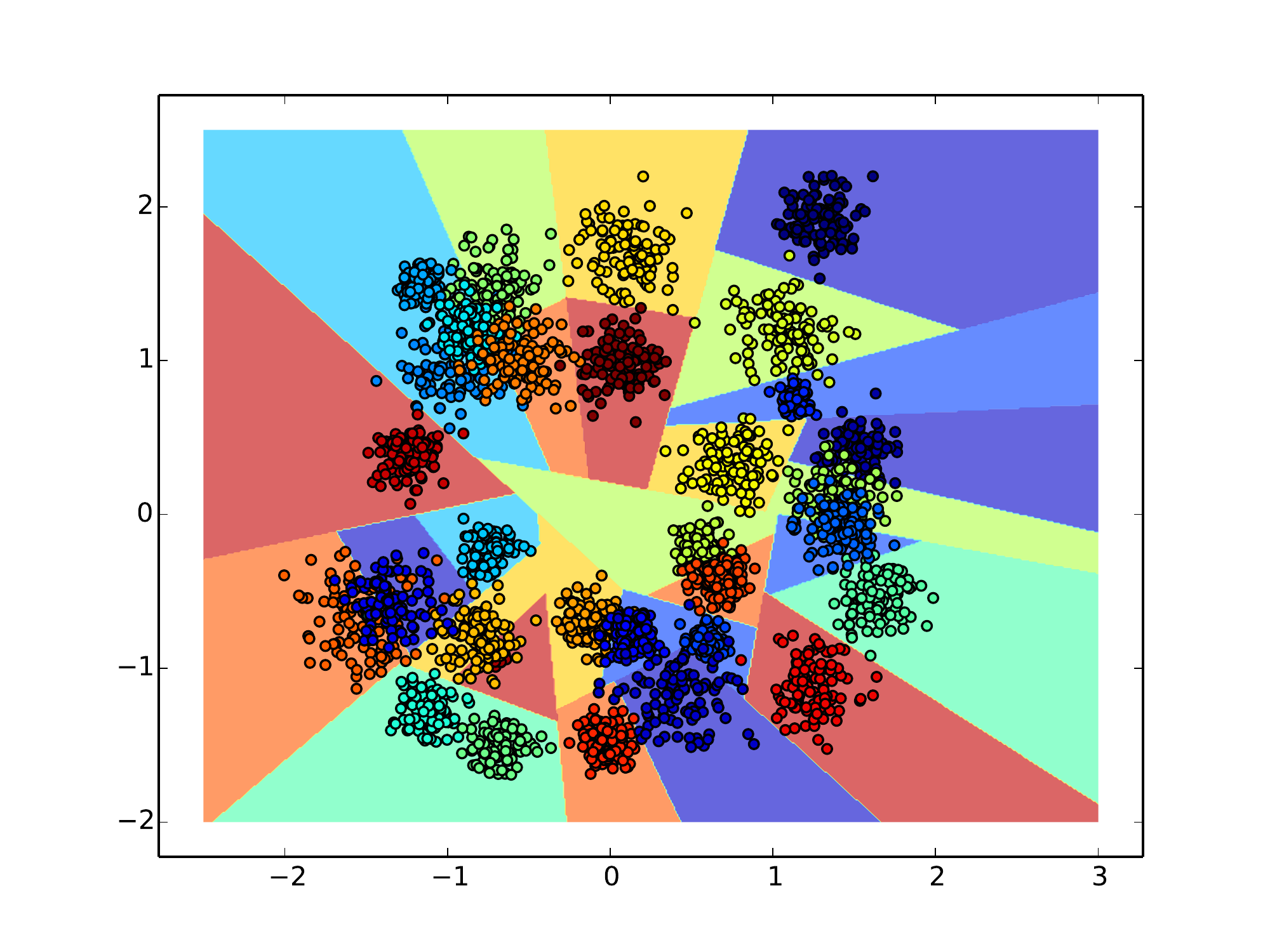}
        \end{minipage}
       
        \caption{Performance (left) on 32 categories and (right) corresponding decision frontiers of RDT: $W$ is the width of the tree (i.e the number of children per node), $D$ is the depth and $L$ is the resulting number of leaves.}
        \label{res3}
\end{figure}

We consider \textbf{simple 2D datasets} composed of categories sampled following a Gaussian distribution. Each category is composed of 100 vectors (50 for train, 50 for test) sampled following $\mathcal{N}(\mu_c,\sigma_c)$ where $\mu_c$ has been uniformly sampled between $\{(-1,-1)$ and $(1,1)\}$ and $\sigma_c$ has been also sampled  - see Figure \ref{res2}-right for the dataset with $C=16$  categories and \ref{res3}-right for the dataset with $C=32$ categories. Two preliminary conclusions can be drawn from the presented results: first, one can see that, when considering a particular architecture, the RDT is able to determine how to allocate the categories in the leaves of the tree: the performance of RDT w.r.t random trees where categories have been randomly sampled in the leaves is clearly better. (ii) The second conclusion is that, when considering a problem with $C$ categories, a good option is to build a tree with $L>C$ leaves since it gives more freedom degrees to the model which will more easily find how to allocate the categories in the leaves. Note that, for the problem with 32 categories, a 84\% accuracy is obtained when using a tree of depth 6: only 6 binary classifiers are used for predicting the category. In comparison to a linear SVM, where the inference complexity is higher than ours ($\mathcal{O}(C)$) our approach performs better. This is mainly due to the ability of our model to learn non-linear decision frontiers. At last, when considering decision trees, one can see that they are sometimes equivalent to RDT. We think that this is mainly due to the small dimension of the input space, and the small number of examples for which decision trees are well adapted.

       

\section{Related Work}
In multi-class classification problems, the classical approach is to train one-versus-all classifiers. It is one of the most efficient technique \citep{Jr2011,Babbar2013a} even with a very large number of classes, but the inference complexity is linear w.r.t the number of possible categories resulting in low-speed prediction algorithms.

  Hierarchical models have been proposed to reduce this complexity. They have been developed for two cases: (i) a first one where a hierarchy of category is already known; in that case, the hierarchy of classifiers is mapped on the hierarchy of classes. (ii) A second approach closer to ours consists in automatically building a hierarchy from the training set. This is usually done in a preliminary step by using for example clustering techniques like spectral clustering on the confusion matrix \citep{Bengio2010}, using probabilistic label tree \citep{Liu2013} or even partitioning optimization \citep{Weston2013}.  
Facing these approaches, RDT has the advantage to learn the hierarchy and the classifier in an integrated step only guided by a unique loss function. The closest work is perhaps \cite{Choromanska2004} which discovers the hierarchy using online learning algorithms, the construction of the tree being made during learning. Other families of  methods have been proposed like error-correcting codes \citep{Dietterich1995a,Schapire1997,Cisse2012}, sparse coding \citep{Zhao2013}  or even using representation learning techniques, representations of categories being obtained by unsupervised models \citep{Weinberger2008,Bengio2010}. 
  
At last, the use of sequential learning models, inspired by reinforcement learning, in the context of classification or regression has been explored recently for different applications like features selection \citep{Dulac-arnold2011} or image classification \citep{dulac, graves}. Our model belongs to this family of approaches.

\section{Conclusion and Perspectives}

We have presented \textit{Reinforced Decision Trees} which is a learning model able to simultaneously learn how to allocate categories in a hierarchy and how to classify inputs. RDTs are sequential decision models where the prediction over one input is made using $\mathcal{O}(\log C)$ classifiers, making this method suitable for problems with large number of categories. Moreover, the method can be easily adapted to any learning problem like regression or ranking, by changing the loss function. RDTs are learned by using a policy gradient-inspired methods. Preliminary results show the effectiveness of this approach. Future work mainly involves real-world experimentation, but also extension of this model to continuous outputs problems. 
\newpage
\small{
\bibliography{library}
}
\end{document}